\documentclass[runningheads]{llncs}

\usepackage{latexsym}
\usepackage{calc}
\usepackage{booktabs}
\usepackage{cite}
\usepackage{graphicx}
\usepackage[table]{xcolor}
\usepackage{amsmath}
\usepackage{hyperref}
\usepackage{amsfonts}
\usepackage{listings}
\usepackage[T1]{fontenc}

\usepackage[utf8]{inputenc}
\usepackage{url}
\usepackage{microtype}
\usepackage{inconsolata}
\usepackage{pgf-pie}

\author{Cedric Möller\inst{1}\orcidID{0000-0001-6700-3482} \and
Ricardo Usbeck\inst{2}\orcidID{0000-0002-0191-7211}}
\authorrunning{C. Möller and R. Usbeck}
\institute{Universität Hamburg, Department of Informatics, Semantic Systems, Germany \email{cedric.moeller@uni-hamburg.de} \and
Leuphana Universität Lüneburg, Institute for Information Systems, Artificial Intelligence and Explainability, Germany \email{ricardo.usbeck@leuphana.de}\\
}

\title{DISCIE - Discriminative Closed Information Extraction}

\begin{document}

\maketitle

\begin{center}
    \fbox{
        \parbox{0.9\linewidth}{
            \textbf{Preprint Notice:} This is a preprint of the following paper:  
            Cedric Möller and Ricardo Usbeck, "DISCIE – Discriminative Closed Information Extraction"  
            in "Lecture Notes in Computer Science", vol. 15232, pp. 23–40 (ESWC 2024 Satellite Events).  
            First online: November 27, 2024.  
            
            The final authenticated version is available at:  
            
            \url{https://link.springer.com/chapter/10.1007/978-3-031-77850-6_2}
        }
    }
\end{center}

\begin{abstract}
This paper introduces a novel method for closed information extraction. The method employs a discriminative approach that incorporates type and entity-specific information to improve relation extraction accuracy, particularly benefiting long-tail relations. Notably, this method demonstrates superior performance compared to state-of-the-art end-to-end generative models. This is especially evident for the problem of large-scale closed information extraction where we are confronted with millions of entities and hundreds of relations. Furthermore, we emphasize the efficiency aspect by leveraging smaller models. In particular, the integration of type-information proves instrumental in achieving performance levels on par with or surpassing those of a larger generative model. This advancement holds promise for more accurate and efficient information extraction techniques.
\end{abstract}

\section{Introduction}

Today, our ability to generate data far surpasses our ability to understand it, particularly when that data is in textual form. As a potential solution, knowledge graphs (KGs), structured representations of data as interconnected nodes and links, offer the promise of making complex information machine-readable and easily interpretable~\cite{DBLP:journals/tnn/JiPCMY22}.

However, the process of automatically transforming unstructured text into a meaningful KG is a significant unsolved problem. It encompasses numerous complex subproblems such as entity recognition, relation extraction and semantic understanding, where each represents a substantial field of study. 

In general, this means that a text is translated to a set of triples. Each triple consists of a subject, a predicate and an object. Each subject and object is an entity while the predicate corresponds to a relation between the two entities.

In this work, we focus on Closed Information Extraction (CIE)~\cite{Josifoski2022}. 
Here, triples are extracted which are grounded in an underlying KG. This means that each of the extracted entities and relations have unique identifiers assigned. 
An example\footnote{Using QIDs and PIDs from \url{www.wikidata.org}. QIDs are the identifiers of entities and PIDs are the identifiers of relations.} is:

\begin{center}
    "Barack Obama was born in Hawaii" $\rightarrow$ \texttt{[Q76, P19, Q782]}
\end{center}

Recent methods like the State-of-the-Art model GenIE interpreted the task as an end-to-end machine translation task where the input is the text and the output are the triples. Generative models are employed to translate text directly to triples. While generative models proved to be very powerful, it is harder to incorporate external information (e.g., the underlying KG) into the generation process~\cite{Josifoski2022}. Hence, the generative model is forced to learn the entire KG during training. As the size of such a KG can be huge, this can inhibit performance. Also, this means such generative methods are not able to use an evolving version of the KG. Furthermore, the sequential nature of the decoding process often to leads lower efficiency, which is critical given the large amount of textual data available today. Lastly, the authors reported a lower performance on long-tail relations.\footnote{Relations rarely occurring.} 

Instead of using generative models, this work employs discriminative models. This means, we first identify salient segments of the input text. Subsequently, external information is introduced to distinguish these segments within a predefined set of classes. The discriminative process encompasses tasks such as recognizing mentions, disambiguating entities, and extracting relations. Also, in many subtasks relevant to the task of end-to-end entity linking are non-generative models still state-of-the-art~\cite{DBLP:conf/emnlp/ShavaraniS23,DBLP:conf/eacl/MaWO23}.
This allows us to tackle three shortcomings of the generative State-of-the-Art model: efficiency, inclusion of external information and performance on long-tail relations. 

We employ lightweight models in each step of our method which gives us a large efficiency boost. While such methods often perform worse than their larger counterparts, we investigate whether the utilization of fine-grained entity type information as external information into relation extraction step can alleviate the performance gap.
Lastly, we explore whether this information has a positive impact on the performance on long-tail relations as well.
The primary emphasis of this paper is centered around enhancements made to the relation extraction component. In terms of mention recognition and entity linking, our approach trains and uses models that have demonstrated high performance.

The contributions of this paper are as follows:
\begin{itemize}
    \item Show that the inclusion of coarse-grained type information is not sufficient;
    \item Show that the inclusion of fine-grained type information has a large impact on relation extraction and hence CIE in general (in particular long-tail relations);
    \item Show that efficient lightweight discriminative models can outperform large-scale generative models when using fine-grained type information.
\end{itemize}

In the following, we will develop such a discriminative method and especially focus on the incorporation of type information into the relation extraction step.\footnote{The code can be found in: \url{https://github.com/semantic-systems/discie}}

\section{Method}
\subsection{Problem definition - Closed Information Extraction}
We define the problem as follows:
Given are a text $t$ and a KG $\mathcal{G}=(V, R, E)$ where $V$ are all entities in the graph, $R$ all relations in the graph and $E\subseteq (V \times R \times V)$ all edges each of which connects two entities via a relation. Each text $t$ contains triples of form $<v, r, w>$ with $v,w \in V$ and $r \in R$. The goal is to extract these triples from text. 
\subsection{Model}
\begin{figure}[!htb]
    \centering
    \includegraphics[width=0.98\textwidth]{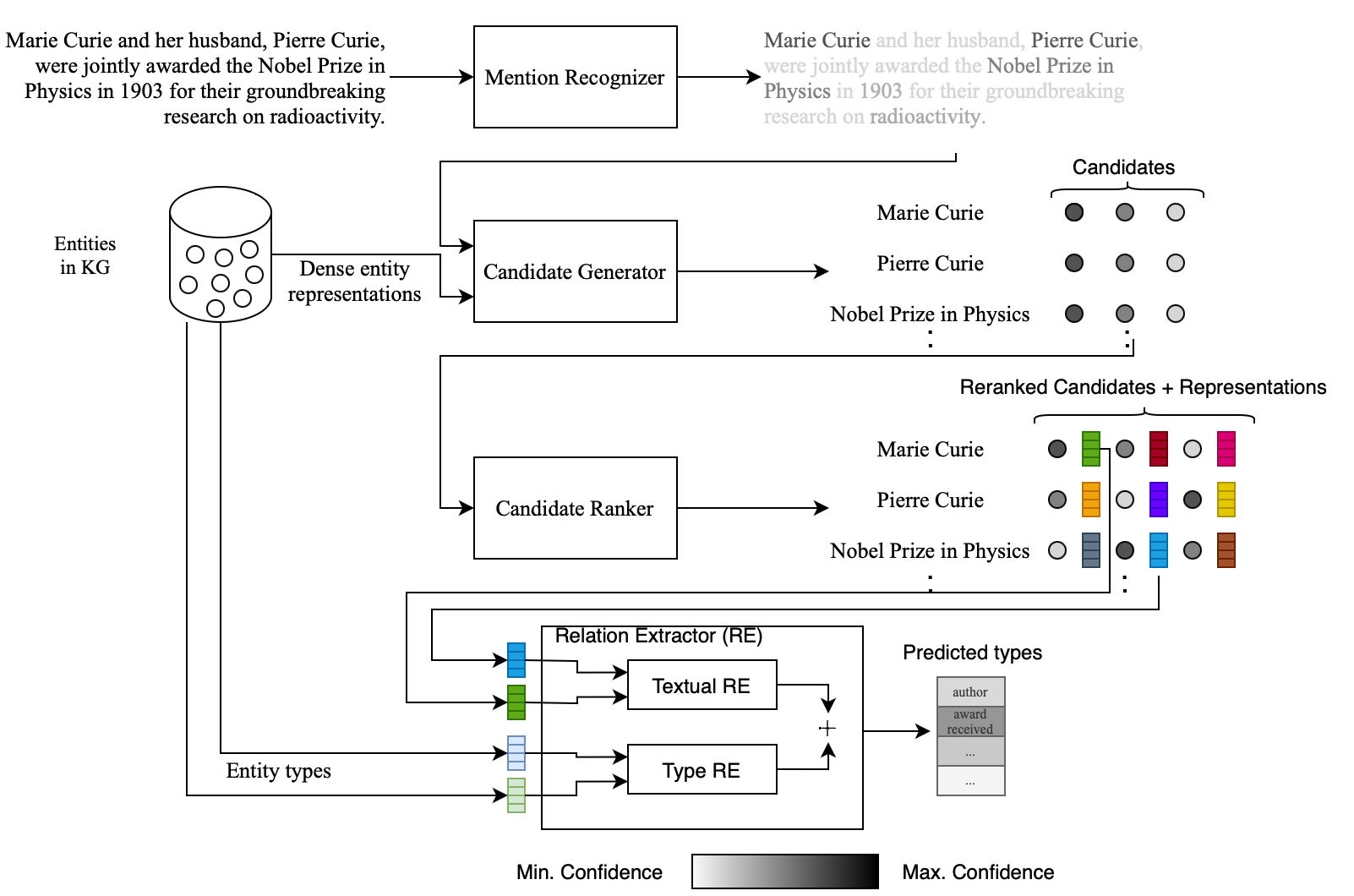}
    \caption{DISCIE - Architecture. The intensity of the colors indicate the scores. Higher intensity resolves to a higher score. The likely outcome would be the triples: [(Q7186:Marie Curie, P166:award received, Q38104:Nobel Prize in Physics), (Q7186:Marie Curie, P26:spouse, Q37463:Pierre Curie)]}
    \label{fig:architecture}
\end{figure}
\subsubsection{Mention Recognizer}

The mention recognizer is an encoder-only model that accepts the tokenized input text $t_1, \dots, t_i , \dots, t_n$. It encodes the whole sequence to get an embedded representation for each token $k_1, \dots, k_i , \dots, k_n$, where $k_i \in \mathbb{R}^d$. Then, each pair of subsequent tokens is combined by concatenation and fed into a linear layer $s_{i,j} = l(k_i \oplus k_j)\in \mathbb{R}$, classifying whether the pair denotes the first and last token of a mention or not. Overall it outputs $\frac{n (n+1)}{2}$ scores for a sequence of length $n$.
All scores surpassing an initial threshold are taken as mention candidates and forwarded to the entity candidate generation module.\footnote{Usually, mention recognition is solved by applying BIO sequence tagging. We trained and evaluated such a method but achieved a lower performance in comparison to the token-pair-based approach described above.}
The model is trained with the binary cross entropy loss function. 

\subsubsection{Entity Candidate Generator}
The entity candidate generator is based on the bi-encoder architecture, more specifically a Siamese network~\cite{chicco2021siamese}. The Siamese network is an encoder-only model.
It encodes the textual mention representation and the textual entity representation into dense representations.
The textual mention representation is of form 
\begin{itemize}
\centering
    \item[] [CLS] \texttt{\{mention\} [CTX\_L] \{context\_left\} [CTX\_R] \{context\_right\}} [SEP] 
\end{itemize}
where \texttt{context\_left} and  \texttt{context\_right} is the text of a certain window size left and right of the identified mention. [CTX\_L] and [CTX\_R] are special tokens denoting the context.
The textual entity representation is of form
\begin{itemize}
\centering
    \item[] [CLS] \texttt{\{label\} [DESC] \{desc\}} [SEP]
\end{itemize}
where \texttt{\{label\}} is the English label of the entity and \texttt{\{desc\}} is the short description text as available in Wikidata via the predicate \texttt{schema:description}.\footnote{This could be replaced with any other KG containing descriptions.}
Both, the mention and entity textual representations, are fed into the same encoder-only model and encoded to retrieve the final mention/entity representation,  which is here taken as the embedded \texttt{[CLS]} token. We denote the embedded mention representations as $b_m$ and the embedded entity representation as $b_c$.
Finally, both representations, are compared via cosine similarity $$\frac{<b_m, b_c>}{\lVert b_m \rVert \lVert b_c \rVert}\in \mathbb{R}$$ where $<\cdot,\cdot>$ denotes the dot product and $\lVert \cdot \rVert$ the euclidean norm. 

The model is trained with the binary cross-entropy loss using in-batch negatives and mined hard-negatives~\cite{Wu2020}.
When doing in-batch negative-based learning, all other entities in the current batch are interpreted as negatives. For mined hard-negatives, all entities are embedded after $\beta$ epochs and for each training example, all $\gamma$ nearest entities are found. All entities not being the ground-truth entity are now taken as negatives. The method returns for each mention $m$ a set of candidates $C_m$. During training, $\beta=1$ and $\gamma=10$.
After training, all entities are embedded and inserted into a vector index for fast retrieval.\footnote{\url{https://faiss.ai}}

\subsubsection{Entity Candidate Ranker}
While the entity candidate generator alone could be used for entity disambiguation, it is usually less accurate. That is why in a subsequent step an entity candidate ranker is used that is less efficient but more accurate. It is applied to the subset of entities retrieved by the previous step. 
The candidate ranker re-ranks all the candidates $C_m$ retrieved for a mention.
It is based on the cross-encoder architecture~\cite{Wu2020}.
It takes the concatenated textual representations of the mention and entity candidate and feeds it into an encoder-only model.  The cross-encoder architecture allows cross-attention between the entity representation and the input text. This usually leads to higher performance than just comparing the bi-encoder representations directly~\cite{Wu2020}.\footnote{This was also observable in our use case.}
Hence, the input is of form:
\begin{itemize}
\centering
    \item[] [CLS] \texttt{\{label\} [DESC] \{desc\} [SEP] \{mention\} [CTX\_L] \{context\_left\} [CTX\_R] \{context\_right\}} [SEP]
\end{itemize}
[DESC] is a special token denoting the entity description.
The embedded \texttt{[CLS]} token is taken (denoted as $b_{m, c}$) and fed into a final linear layer to get a similarity score $$s_{m,c}=h(b_{m, c})\in \mathbb{R}$$.

During training, for each entity mention a set of hard negative entity candidates is sampled by using the entity candidate generator and the vector index.
The model is trained via binary cross entropy loss including all the hard negatives and the positive entity candidate. 

\subsubsection{Relation Extractor}

\paragraph{Textual information.}
The relation extractor accepts a pair of entity mentions. Instead of only focusing on the input text, we incorporate candidate information as well. 
We take each mention $m$ and its highest scoring candidate $c$, and combine both $f(m, c)$.
Then, each $f(m,c)$ is compared to all other $f(m', c')$ where $m \ne m'$.
As the combination of each mention and its candidate $f(m,c)$ we simply use the embedded \texttt{[CLS]} token output by the candidate ranker, so $f(m,c)=b_{m,c}$. 
The predicted relation is scored by first calculating whether a subject-object relationship holds between a pair
$$<l_s(b_{m,c}), l_o(b_{m', c'})> \in \mathbb{R}$$ where $l_s$ and $l_o$ are learnable linear layers.
Then, a score for each potential relation is calculated as 
$$ W_r \left[b_{m,c} + b_{m', c'}\right]  $$ where $W_r\in \mathbb{R}^{|R| \times d}$  is a learnable matrix and $d$ is the dimension of $b_{m,c}$ and $b_{m', c'}$.

Both scores are then combined to get the final relation score $$g[b_{m, c}, b_{m', c'}]  = W_r \left[b_{m,c} + b_{m', c'}\right] + <l_s(b_{m,c}), l_o(b_{m', c'})> \mathbf{1}$$
where $\mathbf{1}\in \mathbb{R}^{|R|}$ is a vector of ones. It holds that $g[b_{m, c}, b_{m', c'}] \in \mathbb{R}^{|R|}$.

\paragraph{Type information.}
Additionally, we also incorporate fine-grained type information into the relation extraction process. This is based on the intuition that certain relations are usually restricted to combinations of certain entity types. To learn these dependencies, we calculate relation classification logits separately from the textual representations just using the type information of each candidate.
Each entity candidate $c$ has a set of types $T_c \subseteq T$ where $T$ is the set of types available in Wikidata. 
Now, we assign each type $t \in T$  a learnable vector $e_t\in \mathbb{R}^{d_\mathcal{T}}$. As an entity might have multiple types, we create a condensed representation of the candidate as $t_c=\frac{1}{|T_c|}\sum_{t \in T_c} e_t$.

Then, we calculate the type-based relation logits by feeding the concatenation of $t_c$ and another candidate $t_{c'}$ into a linear layer: $$h(t_c \oplus t_{c'})\in \mathbb{R}^{|R|}$$

Finally, we sum up the contextual logits and the type logits to get the final logits: $$k(m,c, m', c')=h(t_c \oplus t_{c'}) + g[f(m,c) \oplus f(m', c')]$$

The relation extractor is trained via binary cross-entropy loss. 

\subsubsection{Inference}

First, we retrieve a set of suitable mentions by applying the mention recognizer. After mapping its output to $(0,1)$ by applying the sigmoid function, we retrieve a score $s^m_{i,j}$ for each possible span. Now, all spans surpassing a threshold $\epsilon_m$ are taken as mention candidates. For each such mention, the entity candidate generator is applied to retrieve a set of candidates. Each candidate is reranked by applying the entity candidate reranker. Its scores $s^c_{m,c}$ are again mapped to $(0,1)$. 
The final candidate score is then the average $s= \frac{s^m_{i,j} + s^c_{m,c}}{2}$. If the maximum score of all candidates surpasses a threshold $\epsilon_c$, the candidate and its mention are accepted.
Finally, the relations of each pair of mention-candidate combinations are calculated by using the relation extractor to get the relation scores $s_r$. Each relation score surpassing the final relation threshold $\epsilon_r$ is accepted. 

\begin{table*}[!htb]
    \centering
    \begin{tabular}{lccccc}
    \toprule
        & \multicolumn{4}{c}{REBEL} &\multicolumn{1}{c}{FewRel} \\
       Model & $P$ & $R$ & $F1$ & $F2$ & $R$\\
         \midrule
         SOTA-Pipeline &$43.30 \scriptstyle \pm 0.15$&$41.73 \scriptstyle \pm 0.13$& $42.50 \scriptstyle \pm 0.13$ & - & $17.89 \scriptstyle \pm 0.24$\\
         GenIE &$68.02 \scriptstyle \pm 0.15$&$69.87 \scriptstyle \pm 0.14$& $68.93 \scriptstyle \pm 0.12$ & - & $30.77 \scriptstyle \pm 0.27$\\
         GenIE - PLM &$59.32 \scriptstyle \pm 0.13$&$77.78 \scriptstyle \pm 0.12$& $67.31 \scriptstyle \pm 0.10$ & - & $46.95 \scriptstyle \pm 0.27$\\
         DISCIE (F2 calibrated) & $62.13 \scriptstyle \pm 0.10$  & $\mathbf{81.93 \scriptstyle \pm 0.07}$ & $70.67 \scriptstyle \pm 0.08$ & $\mathbf{77.02 \scriptstyle \pm 0.06}$ & $\mathbf{47.10 \scriptstyle \pm 0.28}$\\
         DISCIE (F1 calibrated)& $\mathbf{77.41 \scriptstyle \pm 0.11}$ & $72.68 \scriptstyle \pm 0.08$ & $\mathbf{74.97 \scriptstyle \pm 0.08}$ & $73.58 \scriptstyle \pm 0.07$ & $34.39 \scriptstyle \pm  0.29$\\
         \bottomrule
    \end{tabular}
    \caption{Results on REBEL and FewRel (Micro)}
    \label{tab:results_cie_full}
\end{table*}

\begin{table*}[!htb]
    \centering
    \begin{tabular}{lcccccc}
        \toprule
        &\multicolumn{3}{c}{GeoNRE} & \multicolumn{3}{c}{WikipediaNRE} \\
           Model & $P$ & $R$ & $F1$ & $P$ & $R$ & $F1$  \\    \midrule
        SOTA-Pipeline &$66.65 \scriptstyle \pm 1.47$ & $66.22 \scriptstyle \pm 1.46$ & $66.43 \scriptstyle \pm 1.45$ & $65.17 \scriptstyle \pm 0.27$ & $54.40 \scriptstyle \pm 0.20$ & $59.30 \scriptstyle \pm 0.21$ \\
        SetGenNet & $86.89 \scriptstyle \pm 0.51$ & $85.31 \scriptstyle \pm 0.47$ & $86.10 \scriptstyle \pm 0.34$ & $82.75 \scriptstyle \pm 0.11$ &$77.55 \scriptstyle \pm 0.27$ &$80.07 \scriptstyle \pm 0.27$ \\
        GenIE & $91.77 \scriptstyle \pm 0.98$ & $\mathbf{93.20 \scriptstyle \pm 0.83}$& $\mathbf{92.48 \scriptstyle \pm 0.88}$ & $91.39 \scriptstyle \pm 0.15$ & $\mathbf{91.58 \scriptstyle \pm 0.15}$ & $91.48 \scriptstyle \pm 0.12$ \\
        DISCIE & $\mathbf{92.4 \scriptstyle \pm 0.90}$ & $87.2 \scriptstyle \pm 1.02$& $89.71 \scriptstyle \pm 0.86$  & $\mathbf{91.57 \scriptstyle \pm 0.16}$& $91.53 \scriptstyle \pm 0.13$ & $\mathbf{91.55 \scriptstyle \pm 0.12}$  \\
        \bottomrule
    \end{tabular}
    \caption{Results on GeoNRE and WikipediaNRE (Micro)}
    \label{tab:results_other_datasets} 

\end{table*}

\section{Evaluation}
For evaluation, we used four different datasets: REBEL, WikipediaNRE, GeoNRE and FewRel.
Here, REBEL~\cite{DBLP:conf/emnlp/CabotN21} is a large-scale dataset while WikipediaNRE, GeoNRE~\cite{Trisedya2019} and FewRel~\cite{DBLP:conf/emnlp/HanZYWYLS18} are of smaller size. In regard to relations, REBEl contains 857 different relations while the other three datasets all contain fewer than 157 relations.
During the evaluation, FewRel is used as a recall-only benchmark dataset as it is not exhaustively annotated~\cite{Josifoski2022}. See Table~\ref{tab:datasets} for information on the datasets.
For the candidate representations, we use the concatenation of the Wikipedia title and the Wikidata description of the entity. The used Wikdata dump is from 2022.
As for type information, we use the fine-grained types as given by the \texttt{P31} relation in Wikidata. Also, we extract for each type all supertypes and consider them as valid types of an entity. Finally, we restrict the set of types to the set as defined by Ayoola et al.~\cite{Ayoola2022} due to them showing great performance utilising them in the task of entity linking.\footnote{930 types are used in total. They were filtered by exploring how useful they are for disambiguating between different entities.}

\begin{table}[]
    \centering
    \begin{tabular}{rrrrrrrrr}
    \toprule
        Dataset & \multicolumn{3}{c}{Examples} & \multicolumn{3}{c}{Triples} & \# entities & \# relations  \\
        & Train & Dev & Test & Train & Dev & Test & & \\
        \midrule
        Rebel & 1,899,331 & 104,960 & 105,516 & 5,147,836 & 284,268 & 284,936 & 1,498,143 & 857 \\
        WikipediaNRE & 223,536 & 980 & 29,619 & 298,489 & 1,317 & 39,678 & 278,204 & 157 \\
        GeoNRE & - & - & 1,000 & - & - & 1,000 & 124 & 11 \\
        FewRel & - & - & 27,650 & - & - & 27,650 & 64,762 & 80 \\
        \bottomrule
    \end{tabular}
    \caption{Statistics of the datasets}
    \label{tab:datasets}
\end{table}

Similar to Josifoski et al.~\cite{Josifoski2022}  we follow two training regimes:
For REBEL and FewRel, we train on the training dataset of REBEL and evaluate on the REBEL test and FewRel test set.
For WikipediaNRE and GeoNRE, we finetune the already REBEL-trained model on the training dataset of WikipediaNRE and then evaluate on the test sets of WikipediaNRE and GeoNRE.\footnote{When evaluating on GeoNRE or WikipediaNRE, we limited the set of available predictable relations and entities to the same set as used in the work by Josifoski et al.~\cite{Josifoski2022}. Therefore, we set prediction scores for out-of-scope relations to $0.0$.}

The thresholds $\epsilon_m$, $\epsilon_c$ and $\epsilon_r$ necessary for inference are tuned on the validation sets of REBEL, respectively WikipediaNRE.

We use \texttt{distilbert-base-cased} for the mention recognizer, and \texttt{all-MiniLM-L12-v2} for the bi-encoder, cross-encoder and relation extractor. While larger models potentially perform better, due the efficiency objective and the fact that we are a small university lab, we rely on such lightweight models.
We train each model for $10$ epochs on two NVIDIA A6000s and select the best-performing model by evaluating on the validation datasets. We use a learning rate of $2 \cdot 10^{-5}$ for all models.

\subsection{CIE Evaluation}
For the closed information extraction task, we compare our trained model, denoted as DISCIE, to the same models as used in the works by Josifoski et al.~\cite{Josifoski2022}. GenIE is the SOTA model by Josifoski et al. utilizing a generative model trained from scratch. GenIE-PLM is the same model but initialized from pre-trained BART~\cite{Lewis2020}. 
SetGenNet~\cite{DBLP:conf/emnlp/SuiW000B21} is a encoder-decoder-based model utilising bi-partite matching for extracting triples. Finally, SOTA-Pipeline is a pipeline-based model by Josifoski et al. relying on a sequence of SOTA models for the tasks of mention recognition, entity linking and relation extraction. For more information on this pipeline, please refer to their paper.~\footnote{ We did not compare to SCICERO~\cite{DBLP:journals/kbs/DessiORBM22} as we were not able to adapt their code to our datasets.}

We report micro/macro precision, recall and F1 for all models as well as F2 for our model.
Micro refers here to calculating the metric over all examples while macro calculates the metrics first for each relation separately and then averages them. 

Similar to Josifoski et al., we report the metrics with a 1-standard-deviation confidence interval constructed from 50 bootstrap samples of the data for all results. 

In Table\ref{tab:results_cie_full},  we show the results on the REBEL and the FewRel datasets.
Our method outperforms the best-performing method GenIE by more than $5$ F1-measure points. 

It can be seen that DISCIE has much higher precision while lacking recall in comparison to GenIE. When tuning the thresholds for F2 instead F1 on the validation dataset~\footnote{Hence putting more emphasis on recall.}, we see that the recall on the subset of data surpasses GenIE while also surpassing it in overall F1. On the recall-only benchmark FewREL, the F2-calibrated DISCIE performs slightly better than GenIE-PLM while the F1-calibrated DISCIE performs much worse. This is the case as the F1-calibrated DISCIE puts more emphasis on precision which leads to a reduced recall.

Table~\ref{tab:results_other_datasets} presents the results for GeoNRE and WikipediaNRE. On GeoNRE, DISCIE performs nearly 3 F1 points worse while on WikipediaNRE it is only slightly better.

On REBEL, macro F1 of DISCIE surpasses the second-best method GenIE by nearly 7 points (see Table~\ref{tab:macro_results_cie_full}) with the F1 calibrated method and by more than 9 points with the F2 calibrated one. This means DISCIE performs more uniformly than GenIE across all relation types. This is especially important due to the large number of relations occurring in the dataset where many are long-tail relations.\footnote{They occur only rarely in the training data.}

\begin{table*}[!htb]
    \centering
    \begin{tabular}{lcccc}
    \toprule
       Model & $P_{\text{Macro}}$ & $R_{\text{Macro}}$ & $F1_{\text{Macro}}$ & $F2_{\text{Macro}}$ \\
         \midrule
         SOTA-Pipeline &$12.20 \scriptstyle \pm 0.35$&$10.44 \scriptstyle \pm 0.22$& $9.48 \scriptstyle \pm 0.21$ & -\\
         GenIE &$33.90 \scriptstyle \pm 0.73$&$30.48 \scriptstyle \pm 0.65$& $30.46 \scriptstyle \pm 0.62$ & -\\
         DISCIE (F2 calibrated) & $35.84 \scriptstyle \pm 0.59$ & $\mathbf{43.99 \scriptstyle \pm 0.61}$ & $\mathbf{39.50 \scriptstyle \pm 0.56}$ & $\mathbf{42.08 \scriptstyle \pm 0.57}$ \\
         DISCIE (F1 calibrated)& $\mathbf{44.05 \scriptstyle \pm 0.84}$ & $42.29 \scriptstyle \pm 0.62$ & $37.27 \scriptstyle \pm 0.67$ & $34.11 \scriptstyle \pm 0.63$\\
         \bottomrule
    \end{tabular}
    \caption{Results on REBEL (Macro)}
    \label{tab:macro_results_cie_full}
\end{table*}

On WikipediaNRE and GeoNRE, while DISCIE sometimes underperformed or only matched the performance of GenIE on micro metrics, we see that in regard to macro metrics, it outperforms GenIE  (see Table~\ref{tab:macro_results_other_datasets}). On WikipediaNRE it outperforms GenIE by 8 points on F1. On GeoNRE, DISCIE surpasses GenIE by more than 3 F1 points. DISCIE is therefore also performing more uniformly on those datasets.

\begin{table*}[!htb]
    \centering
    \begin{tabular}{lcccccc}
    \toprule
    &\multicolumn{3}{c}{GeoNRE} & \multicolumn{3}{c}{WikipediaNRE} \\
       Model & $P_{\text{Macro}}$ & $R_{\text{Macro}}$ & $F1_{\text{Macro}}$ & $P_{\text{Macro}}$ & $R_{\text{Macro}}$ & $F1_{\text{Macro}}$ \\
         \midrule
         SOTA-Pipeline &$38.67 \scriptstyle \pm 5.72$ & $34.49 \scriptstyle \pm 5.99$ & $35.14 \scriptstyle \pm 5.09$ & $24.12 \scriptstyle \pm 1.46$ & $16.55 \scriptstyle \pm 1.00$ & $17.76 \scriptstyle \pm 1.01$  \\
         GenIE & $\mathbf{75.77 \scriptstyle \pm 7.80}$ & $71.60 \scriptstyle \pm 7.95$ & $72.59 \scriptstyle \pm 7.32$ & $52.55 \scriptstyle \pm 2.12$ &$45.95 \scriptstyle \pm 1.67$ & $47.08 \scriptstyle \pm 1.68$  \\
        DISCIE & $73.65 \scriptstyle \pm 6.61$ & $\mathbf{76.72 \scriptstyle \pm 6.54}$& $\mathbf{75.05 \scriptstyle \pm 6.01}$ & $\mathbf{53.76 \scriptstyle \pm 2.14}$& $\mathbf{51.80 \scriptstyle \pm 2.05}$ & $\mathbf{52.75 \scriptstyle \pm 1.90}$  \\
        \bottomrule
    \end{tabular}
    \caption{Results on GeoNRE and WikipediaNRE (Macro)}
    \label{tab:macro_results_other_datasets}
\end{table*}

\begin{figure}[!htb]
    \centering
    \includegraphics[width=0.95\textwidth]{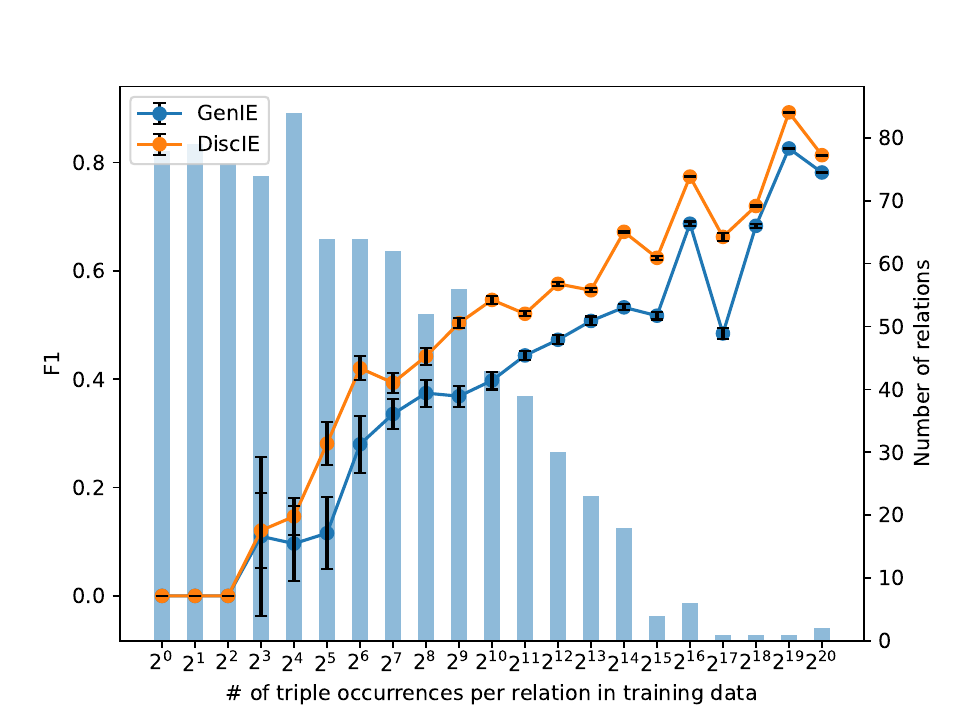}
    \caption{F1 for GenIE and DISCIE over REBEL plotted for buckets of relations; each bucket contains all relations occurring a specific number of times in the training data. Each blue bar shows the number of relations occurring in the \# of triples as given by the x-axis (see right vertical axis). }
    \label{fig:bucket_plot}
\end{figure}

Figure~\ref{fig:bucket_plot} compares the F1 for all relations separated by their number of occurrences in the training data on REBEL. As can be seen, the performance of DISCIE is surpassing the performance of GenIE consistently. For long-tail entities with an occurrence count between $16$ and $64$ ($2^4$ and $2^6$), the performance sometimes nearly doubles.  

\subsection{Ablation}

To identify what aspects of the relation extractor contributed the most to the performance, we conducted an ablation study on REBEL. 
Table~\ref{tab:ablation_relation} compares regular DISCIE to DISCIE without any type information (w/o types), DISCIE without candidate descriptions (w/o desc.) and DISCIE with coarse-grained types (w/ coarse).

Excluding type information (w/o types) leads to a large decrease in performance. Especially the precision decreases by many points. Therefore, implicit KG information given by type information helps the model to more precisely decide on the right relation while filtering out relations not compatible with the types provided by the entities.

Not using candidate information (w/o candidate description) and only relying on the textual information at hand leads to a slight decrease in performance by around $0.5$ F1 points. Hence, the information contained in the description is not fully replaced by the available type information. 

\begin{table}[!htb]
    \centering
    \begin{tabular}{cccc}
    \toprule
        Model & P & R & F1 \\
        \midrule
        DISCIE w/o types& $62.41 \scriptstyle \pm 0.07$ & $69.08 \scriptstyle \pm 0.08$ & $65.58 \scriptstyle \pm 0.06$ \\
        DISCIE w/o desc.& $76.82 \scriptstyle \pm 0.11$ & $72.14 \scriptstyle \pm 0.07$& $74.41 \scriptstyle \pm 0.07$ \\
        
        DISCIE w/o text& $59.75 \scriptstyle \pm 0.24$ & $35.87 \scriptstyle \pm 0.09$ & $44.83 \scriptstyle \pm0.10$\\
        DISCIE w/ coarse & $68.32 \scriptstyle \pm 0.08$& $68.31 \scriptstyle \pm 0.08$& $68.32 \scriptstyle \pm 0.06$\\
         DISCIE &  $\mathbf{77.41 \scriptstyle \pm 0.11}$ & $\mathbf{72.68 \scriptstyle \pm 0.08}$ & $\mathbf{74.97 \scriptstyle \pm 0.08}$ \\
         
         \bottomrule
         
    \end{tabular}
    \caption{Ablation study of the relation extractor evaluated over REBEL dataset (w/o types: relation extractor does not use type information, w/o desc.: relation extractor does not use candidate descriptions, w/o text: relation extractor does neither use candidate descriptions nor the input text, w/ coarse: regular relation extractor but coarse-grained types are used) }
    \label{tab:ablation_relation}
\end{table}

Only using type information (w/o text) and not relying on any textual or candidate description information leads to the largest decrease in performance. The model is still able to often predict the correct relation by just using the available type information. Some combination of types therefore strongly predict the occurrence of certain relations in the text. However, the task is not trivial and therefore textual information at hand is still a necessity.

Lastly, replacing the fine-grained types with coarse-grained types of form \texttt{PER}, \texttt{ORG}, \texttt{LOC}, \texttt{MISC} (w/ coarse) leads to an increase in performance in comparison to not using type information at all. Nevertheless, using fine-grained types increases the performance much more.  

\subsection{Efficiency}
We evaluate the efficiency via the GeoNRE dataset by running GenIE and DISCIE three times on its evaluation dataset.\footnote{GenIE takes a long time to evaluate on the other datasets on a single GPU. Therefore we opted for only running the efficiency tests on the smallest dataset. While the average speed differs between the datasets, DISCIE was considerably faster for all of them. }
Due to the length differences of the examples, the average number of seconds per 1000 examples can vary between datasets. 
In the GeoNRE dataset, DISCIE is approximately 27 times as fast as GenIE while outperforming it or matching it on several benchmarks (see Table~\ref{tab:efficiency}). 
\begin{table}[!htb]
    \centering
    \begin{tabular}{cc}
    \toprule
        Model& Seconds/1000 Examples  \\
        \midrule
         DISCIE&  $21.17 \scriptstyle \pm 0.62$  \\
         GenIE &  $571.95  \scriptstyle \pm 7.08$\\
         \bottomrule
    \end{tabular}
    \caption{Efficiency on GeoNRE dataset run on a single NVIDIA A6000 GPU}
    \label{tab:efficiency}
\end{table}

\subsection{Error Analysis}

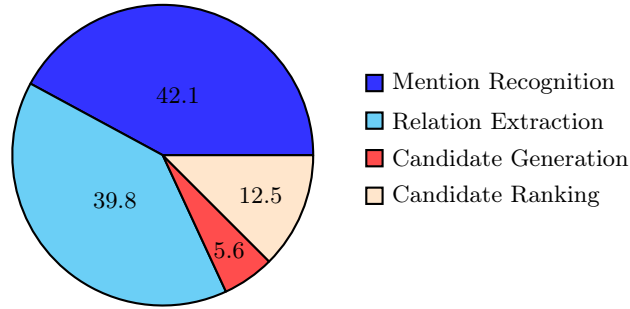
\begin{figure}[!htb]
\centering
\begin{tikzpicture}
\pie[radius=2.0,sum=auto,text=legend,color={blue!80,cyan!50,red!70,orange!20}]{42.1/Mention Recognition,
    39.8/Relation Extraction,
    5.6/Candidate Generation,
    12.5/Candidate Ranking}
\end{tikzpicture}
    \caption{Error distribution over all components on REBEL}
    \label{fig:pie_rebel}
\end{figure}

Figure~\ref{fig:pie_rebel} shows what components amount to what percentage of error on REBEL. As can be seen, is the candidate generation the least prone to errors. Usually, the correct candidate is in the generated candidate set. Candidate ranking is more prone to errors than the candidate generation. Sometimes, the wrong candidate is ranked the highest. The components that contribute the most to the errors are the relation extraction and mention recognition.

Additionally, we compare the results for three different examples for both GenIE and DISCIE in Table~\ref{tab:errors}. Example 1 shows that DISCIE often performs better when focusing on long-tail relations. Here, DISCIE predicts both the relations \texttt{employer} and \texttt{musical conductor} while GenIE only predicts \texttt{member of}. While \texttt{member of} is close to \texttt{employer}, \texttt{employer} is more specific. On the other hand, \texttt{musical conductor}, a long-tail relation, is not predicted by GenIE while DISCIE can predict these.
In Example 2 GenIE predicted the correct relation \texttt{employer} while DISCIE mistakenly predicted \texttt{educated at}. Here, \texttt{educated at} can also be seen as a fitting relation but it was just not labeled. Note, that this a common occurrence with GenIE, namely triples are predicted that are reasonable but not labeled. 
Lastly, Example 3 shows a case where both GenIE and DISCIE fail. Here,  both methods generate more triples than necessary. Most of them describe implicit relations. A notable exception is the triple \texttt{(Spain, capital, Madrid)} that is predicted by both models. This relation is not stated in the input text but both models likely just predict it due to it often being seen during training.

\newcommand{\width}{0.5\textwidth-0.9cm-0.15cm}

\begin{table*}[!htb]
\centering
\begin{tabular}{p{1.1cm}p{\width}@{\hspace{0.2cm}}p{\width}}
\toprule
\textbf{Method} & \textbf{DISCIE} & \textbf{GenIE}  \\ \midrule
 Ex. 1& \multicolumn{2}{p{\textwidth - 1.8cm}}{\cellcolor{gray!25} In 2009, Vásquez was named a Gustavo Dudamel conducting fellow with the Los Angeles Philharmonic. }\\
 Result 1& \textbf{(Gustavo Dudamel, employer, Los Angeles Philharmonic), (Los Angeles Philharmonic, musical conductor, Gustavo Dudamel)} & (Gustavo Dudamel, member of, Los Angeles Philharmonic) \\
 \midrule
  Ex. 2& \multicolumn{2}{p{\textwidth - 1.8cm}}{\cellcolor{gray!25} She earned her Ph.D in mathematics from the University of Illinois at Urbana–Champaign in 1919 under the supervision of Arthur Byron Coble.  }\\
 Result 2& (Arthur Byron Coble, educated at, University of Illinois at Urbana–Champaign) & \textbf{(Arthur Byron Coble, employer, University of Illinois at Urbana–Champaign)} \\
 \midrule
  Ex. 3& \multicolumn{2}{p{\textwidth - 1.8cm}}{\cellcolor{gray!25} The Santiago Bernabéu Stadium (, ) is a football stadium in Madrid, Spain.}\\
 Result 3&  \textbf{(Santiago Bernabéu Stadium, sport, Association football), (Santiago Bernabéu Stadium, located in the administrative territorial entity, Madrid)}, (Santiago Bernabéu Stadium, country, Spain), (Madrid, country, Spain),(Spain, capital, Madrid), (Santiago Bernabéu Stadium, instance of, stadium)
 & \textbf{(Santiago Bernabéu Stadium, sport, Association football), (Santiago Bernabéu Stadium, located in the administrative territorial entity, Madrid)}, (Santiago Bernabéu Stadium, country, Spain), (Madrid, country, Spain),(Spain, capital, Madrid)
 \\
 \bottomrule

\end{tabular}
\caption{Comparison of the performance of DISCIE and GenIE on three different examples. Ground-truth triples are shown in bold }
\label{tab:errors}
\end{table*}

\section{Related work}

Entity Linking has a long history of research~\cite{DBLP:journals/semweb/MollerLU22}. Recent methods can be categorized into two types. First, discriminative methods that are based on the bi-encoder / cross-encoder pairing~\cite{Wu2020,Logeswaran2019,Ayoola2022}. Both encoders are commonly BERT-like models. The bi-encoder encodes the description of each entity and matches it to the text by using an approximate nearest neighbor search. This is important as the next step, the cross-encoding, is expensive. Here, those neighbors are reranked by applying a cross-encoder to the concatenation of both, the input text and the entity description. The highest-ranked entity is then the final linked one. 
In the past, type information was used in several works in the entity linking domain. Incorporating it lead to a large increase  in performance~\cite{raiman2018deeptype,raiman2022deeptype,Ayoola2022}. In contrast to that, we do not employ type information during entity linking but during relation extraction.
Another type of entity linker is based on generative models~\cite{Cao2021,Cao2022}. Here, instead of using some external description of an entity, the whole model memorizes the knowledge graph (KG) during training. The linked entity is then directly generated by the model. Such methods skip the problem of mining negatives which are crucial for a good performance of bi-encoder-based methods~\cite{Cao2021}. 

Relation extraction methods usually assume that the entities in the input text are already identified. The task is then to classify whether a relation between two entities is expressed in the text and if it is, what relation holds.  
Recent methods rely either on CNN~\cite{Santos2015,Zeng2014,Nguyen2015}, RNN~\cite{Ni2019,Miwa2016} or transformer networks~\cite{Zhong2021,Soares2019}. Also, generative models are applied, usually by extracting entities and relations jointly~\cite{Paolini2021,Zhang2020} but also methods solely focusing on relation extraction (RE) exist~\cite{DBLP:conf/emnlp/CabotN21, Ni2022}.
In contrast to DISCIE, these methods generally focus on a small number of relations and do not consider the entity linking task. 
Zhang et al.~\cite{DBLP:journals/corr/abs-2206-05123} include fine-grained information into a generative joint entity and relation extraction method. But in contrast to us they only focus on entity extraction and not entity linking. Furthermore, they only incorporate a single type per entity.  

There exist two directions of research related to closed information extraction. First, pipeline-based approaches. For that, initially, the entities in the text were recognized, then the relations between the entities were identified and finally, relations and entities are linked to the KG~\cite{Angeli2015,Galarraga2014,Chaganty2017}. While the modularity of pipeline-based approaches makes it possible to simply exchange some modules with a better one, they suffer from error propagation. 
To combat that, recent methods focus on tackling the problem end-to-end~\cite{Trisedya2019,Sui2021,Liu2018}. Here, each step of the pipeline is jointly executed at once. This enables the models to have interaction between the entity recognition, relation extraction and entity linking process.  Lately, generative models like BART, T5, GPT-4 became more popular~\cite{DBLP:conf/acl/LewisLGGMLSZ20,DBLP:journals/jmlr/RaffelSRLNMZLL20,DBLP:journals/corr/abs-2303-08774}. Usually, the tasks are here simply modeled as the translation of text to text.  In 2022, Josifoski et al.~\cite{Josifoski2022, DBLP:conf/emnlp/JosifoskiSP023} applied such a generative model to the CIE task reaching SOTA. Furthermore, they are the first two evaluate the CIE task on a large dataset with hundreds of relations and millions of entities. 
Our method is the first discriminative approach focusing on the large-scale closed-information extraction task. In contrast to GenIE by Josifoski et al., we do not rely on a generative model, but a discriminative one. Furthermore, instead of performing relation extraction solely on the textual data, we incorporate the entity candidate information in form of their descriptions and types. Both features prove to be especially valuable when doing the relation extraction task on datasets with a large number of relations.

\section{Conclusion and Future work}
In this work, we showed that including fine-grained type information into a discriminative closed information extraction method leads to a large improvement. By using the type information, the model can learn the implicit ontological information contained in the underlying KG. It especially leads to an \textbf{increased performance on long-tail relations}. Furthermore, due to the reliance of DISCIE on only smaller language models, it can deliver great performance while being much \textbf{more efficient}. This allows our model to match or even surpass the performance of larger end-to-end CIE information models while being much faster.

A generative model such as GenIE can be trained on the closed information extraction task without having access to the entity mention positions. In contrast to that, our training setup relies on them. In future work, we want to investigate whether the model can be modified to skip the mention recognition.
Furthermore, the inference procedure is currently performed in a greedy way. We suspect that globally optimizing the disambiguation graph can lead to an increase in performance, which we also want to pursue further in the future.
Also, incorporating the type information into the entity linking module might lead to improvement. Finally, analysing which types have a bigger impact on performance is worth exploring. 

\section*{Limitations}
SynthIE~\cite{DBLP:conf/emnlp/JosifoskiSP023} showed that the REBEL dataset suffers from some qualitative problems such as false negatives. We did not compare our method against a larger generative model, such as LLama~\cite{DBLP:journals/corr/abs-2302-13971}. Although an adapter-fine-tuned variant of such a large language model might potentially outperform our method, it would require a significantly larger parameter count and be less efficient. Our objective was to demonstrate that substantial performance improvements can be achieved even with a smaller parameter count and some external data.

\paragraph*{Supplemental Material Statement:}
Source code for our System is available from: \url{https://github.com/semantic-systems/discie}

\section*{Acknowledgments}
This project was supported by the Hub of Computing and Data Science (HCDS) of Hamburg University within the Cross-Disciplinary Lab program. Additionally, support was provided by the Ministry of Research and Education within the SifoLIFE project "RESCUE-MATE: Dynamische Lageerstellung und Unterstützung für Rettungskräfte in komplexen Krisensituationen mittels Datenfusion und intelligenten Drohnenschwärmen" (FKZ 13N16836).

\bibliographystyle{splncs04}
\bibliography{main}

\end{document}